\documentclass[letterpaper, 10 pt,conference]{ieeeconf}

\IEEEoverridecommandlockouts
\usepackage{tablefootnote}
\usepackage{array}
\usepackage{cite} 
\usepackage{epsfig}
\usepackage{chngcntr}
\usepackage{amssymb}
\usepackage{amsmath}
\usepackage{graphicx}
\usepackage{verbatim}
\usepackage{float}
\usepackage{booktabs}
\usepackage{multirow}
\usepackage{tabularx}
\usepackage{threeparttable}
\usepackage{svg}
\usepackage{xcolor}
\usepackage{dsfont}
\usepackage[lined,ruled,commentsnumbered, linesnumbered]{algorithm2e}
\usepackage[text={7in,9.5in},centering]{geometry}

\newcommand{\xmobility}{\textsc{X-Mobility}}

\title{\Large\bf
\xmobility{}: End-To-End Generalizable Navigation via World Modeling
}

\author{Wei Liu$^{1*}$, Huihua Zhao$^{1*}$, Chenran Li$^{1,2}$, Joydeep Biswas$^{1,3}$, Billy Okal$^{1}$, Pulkit Goyal$^{1}$,\\Yan Chang$^{1}$ and Soha Pouya$^{1}$
\thanks{$^{*}$ Equal contribution}
\thanks{$^{1}$ Wei Liu, Huihua Zhao, Chenran Li, Joydeep Biswas, Billy Okal, Pulkit Goyal, Yan Chang, and Soha Pouya are with NVIDIA, Santa Clara, California, USA {\tt\small\{liuw, huihuaz, chenranl, jbiswas, bokal, pulkitg, yachang, spouya\}@nvidia.com}}
\thanks{$^{2}$ Chenran Li is also with UC Berkeley, Berkeley, California, USA. The work is conducted during Chenran Li's internship at NVIDIA \tt\small{chenran\_li@berkeley.edu}}
\thanks{$^{3}$ Joydeep Biswas is also with UT Austin, Austin, Texas, USA. \tt\small{joydeepb@cs.utexas.edu}}
}

\begin{document}

\maketitle
\pagestyle{empty}

\begin{abstract}

General-purpose navigation in challenging environments remains a significant problem in robotics, with current state-of-the-art approaches facing myriad limitations. Classical approaches struggle with cluttered settings and require extensive tuning, while learning-based methods face difficulties generalizing to out-of-distribution environments. This paper introduces \xmobility{}, an end-to-end generalizable navigation model that overcomes existing challenges by leveraging three key ideas. First, \xmobility{} employs an auto-regressive world modeling architecture with a latent state space to capture world dynamics. Second, a diverse set of multi-head decoders enables the model to learn a rich state representation that correlates strongly with effective navigation skills. Third, by decoupling world modeling from action policy, our architecture can train effectively on a variety of data sources, both with and without expert policies—off-policy data allows the model to learn world dynamics, while on-policy data with supervisory control enables optimal action policy learning. Through extensive experiments, we demonstrate that \xmobility{} not only generalizes effectively but also surpasses current state-of-the-art navigation approaches. Additionally, \xmobility{} also achieves zero-shot Sim2Real transferability and shows strong potential for cross-embodiment generalization.

Project page: https://nvlabs.github.io/X-MOBILITY

\end{abstract}

\section{Introduction}
Developing a versatile navigation stack for robots is challenging due to the need for adaptability across diverse environments and robot platforms. Such a system must adjust to varying operational constraints while maintaining high efficiency, particularly when operating with limited onboard computational resources. Classical approaches \cite{macenski2020marathon2, liu2015autonomous} often rely on modular architectures, integrating separate navigation components to meet specific needs. Although this modular design offers some generalization, it typically requires complex configurations and extensive tuning for each new deployment \cite{mavrogiannis2023core}. Moreover, these methods often struggle to manage uncertainties in both the environment and the robot's state, which can compromise system robustness. To address this limitation, the Partially Observable Markov Decision Process (POMDP) \cite{spaan2012partially} framework has been proposed, which models probabilistic belief states and solve decision-making problems by interleaving observation and action. However, POMDPs demand significant system modeling effort \cite{liu2015situation}, limiting scalability, and their computational intensity poses challenges for real-time deployment on edge devices \cite{somani2013despot}.

In response to these challenges, end-to-end learning-based navigation stacks have emerged as a promising alternative \cite{mirowski2016learning, Doshi24-crossformer, sridhar2023nomad}. These systems leverage experience to optimize navigation performance and generalize across various tasks. Toward this end, reinforcement learning has been widely explored in robotics, but its application to navigation has been limited by issues such as low sampling efficiency in large-scale environments and difficulties in scaling simulations \cite{zhu2021deep}. Imitation Learning (IL) presents an alternative by enabling robots to learn directly from demonstrations, which has been particularly successful in fields where large-scale datasets with teacher policies are readily available \cite{qin2021deep}. However, IL often encounters the problem of covariate shift, where performance deteriorates in scenarios outside the training data distribution \cite{zare2024survey}. To mitigate this, methods such as DAgger \cite{ross2011reduction} have been proposed, which iteratively expand the training dataset by incorporating the model's errors and the expert's corrections. Additionally, recent advancements in world modeling \cite{ha2018world}, which aim to learn world dynamics in both in-distribution and out-of-distribution environments, have shown promise in enhancing robustness, particularly in autonomous driving \cite{hu2022model, jia2023adriver, wang2023drivewm}. However, a key challenge in applying these approaches to robot navigation is the scarcity of large-scale datasets, especially those needed for imitation learning.

\begin{figure}[t]
\begin{center}
\centering\includegraphics[width=3.4in]{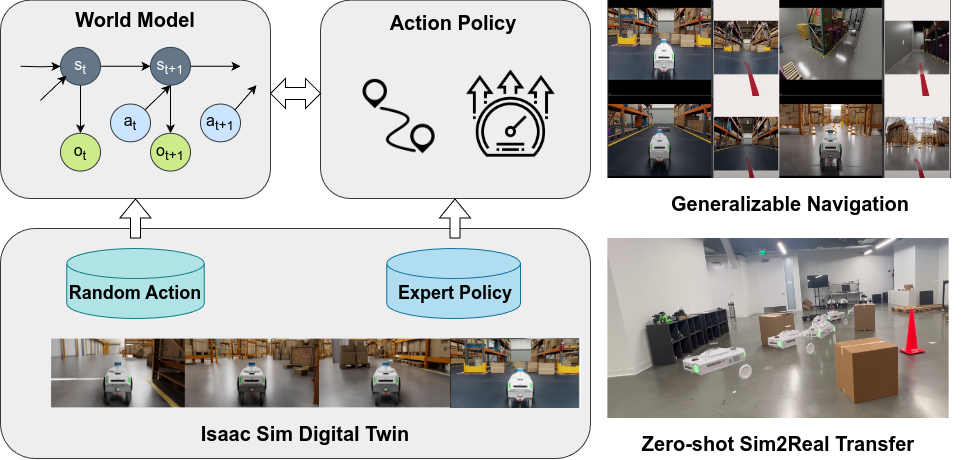}
\caption{\xmobility{}: an end-to-end world model based navigation stack featuring a multi-stage training pipeline using photorealistic synthetic datasets, demonstrating generalizability across out-of-distribution scenarios and zero-shot Sim2Real transferability.}
\label{fig:x_mobility}
\end{center}
\end{figure}

Recognizing these challenges, this paper introduces \xmobility{}, an end-to-end navigation model designed to generalize across various mobility applications (see Fig. \ref{fig:x_mobility}). Inspired by world modeling, \xmobility{} employs a lightweight auto-regressive learning architecture to develop a rich latent representation space that probabilistically captures the world state and its dynamics. This representation space is effectively trained and strongly correlated with robust navigation skills through a set of diverse multi-task decoders. To address data scarcity, \xmobility{} decouples world modeling from action policy imitation, allowing it to train from a variety of data sources, both with and without supervisory control inputs: off-policy data sources enable the model to learn world dynamics, while on-policy sources with supervisory control facilitate the action policy learning.

With this architectural design, \xmobility{} is trained through a multi-stage pipeline that leverages NVIDIA’s Isaac Sim to generate large-scale, photorealistic synthetic datasets featuring diverse scenes and action policies. Extensive experiments demonstrate that \xmobility{} can consistently outperform current state-of-the-art methods and exhibits zero-shot Sim2Real transferability, showing good potential for generalization across different robot embodiments. The key contributions of this paper are:
\begin{itemize}
\item Introduction of an end-to-end, world model-based navigation model that outperforms state-of-the-art methods and demonstrates zero-shot mobility in challenging environments not present in the training data.
\item Design of a model architecture with decoupled world modeling and action policy networks, enabling efficient training from a wide range of data sources, thus addressing data scarcity challenges in robot navigation.
\item Validation of the feasibility of training the navigation model with photorealistic synthetic datasets, achieving zero-shot Sim2Real transferability.
\end{itemize}

The remainder of this paper is structured as follows: Section \ref{section:x-mobility} presents the \xmobility{} model architecture and its component design. Section \ref{section:exp_setting} details the experimental setup, and the results are discussed in Section \ref{section:results}. Section \ref{section:discussion} discusses the Sim2Real transfer and cross-embodiment generalization, and the paper concludes in Section \ref{section:future_work}.

\begin{figure*}[ht]
\begin{center}
\centering\includegraphics[width=6.8in]{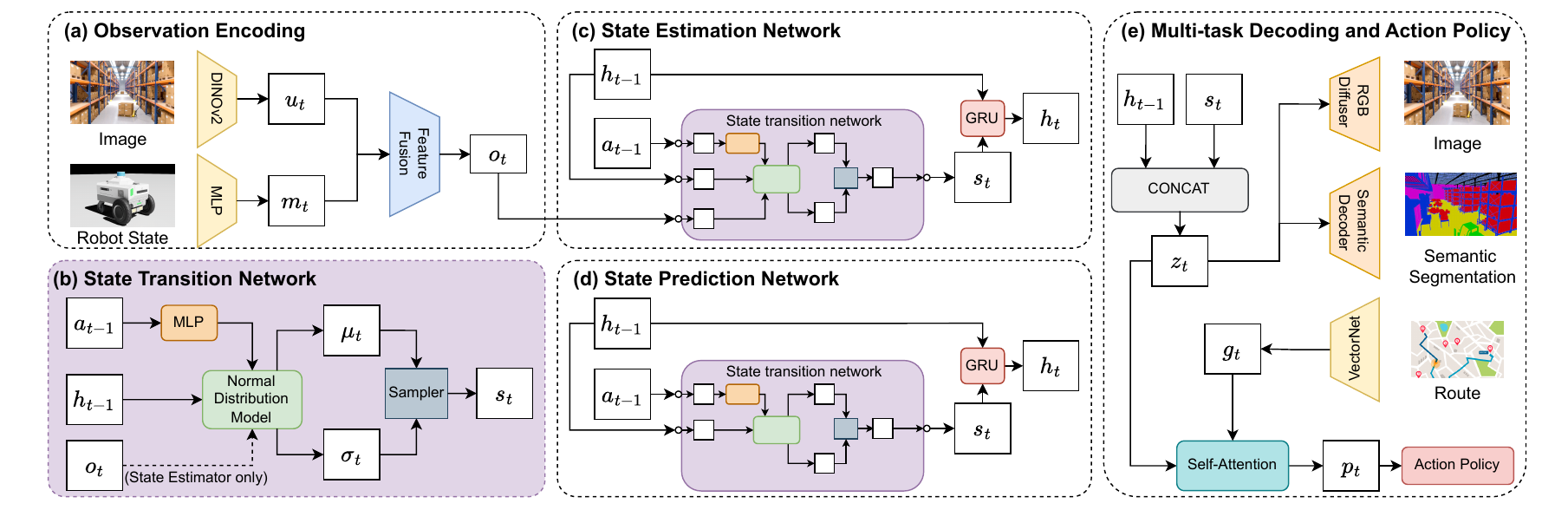}
\caption{Model architecture: i) The state estimator in (c) and state predictor in (d) are designed to capture world dynamics; ii) Along with the multi-task decoders in (e), they generate a rich latent space representation for action policy learning, which takes the latent state and route embedding as inputs.}
\label{fig:model_arch}
\end{center}
\end{figure*}

\section{\xmobility{}}\label{section:x-mobility}
\subsection{Problem Formulation}
The general-purpose robot navigation problem can be formulated as a POMDP, defined by the tuple $\{\mathcal{S}, \mathcal{A}, \mathcal{O}, T, O, R, \gamma\}$ consisting of the state space $\mathcal{S}$,  action space $\mathcal{A}$, the observation space $\mathcal{O}$, transition function $T(s',s,a)=\mathrm{Pr}(s'|s,a)$, and observation function $O(o,s',a)=\mathrm{Pr}(o|s',a)$. The reward function $\mathrm{R}(s,a)$ defines the reward for performing action $a$ in state $s$, and $\gamma \in [0, 1)$ is the discount factor. The solution to the POMDP is an optimal policy $\pi^{*}$ that maximizes the expected accumulated reward, $E\left(\sum_{t=0}^{\infty}\gamma^{t}\mathrm{R}(a_{t},s_{t})\right)$, where $s_{t}$ and $a_{t}$ represent the agent's state and action at time $t$.

\xmobility{} addresses the challenge of generalizable navigation by solving the POMDP within a learned state representation space, which includes a state estimation network, a state prediction network, and an action policy network as in Fig. \ref{fig:model_arch}. The state estimation and prediction networks focus on learning world dynamics, while the action policy network aims to solve the POMDP by imitating the teacher policy, assuming it closely approximates the optimal policy $\pi^{*}$.

\subsection{Belief Representation In Learned State Space}
In classical robot navigation, the state space is often manually defined, such as $\mathds{SE}(2)$ for the robot's location. Previous efforts have enriched state representations by incorporating semantic features like tracked obstacles \cite{crespo2020semantic} or terrain characteristics \cite{xiao2021learning}, but defining a sufficiently expressive and computationally tractable state space for generalizable navigation remains a challenge.

Instead of manually defining the state space, \xmobility{} uses a learned state representation to capture all relevant navigation features. To handle the partial observability inherent in the state space, \xmobility{} maintains a belief over possible states, approximated using a parameterized Normal distribution $s \sim \mathcal{N}(\mu_\theta, \sigma_\theta)$, where $\mu_\theta$ and $\sigma_\theta$ denote the mean and covariance, respectively. These statistics are continuously estimated and updated by the state estimation network, ensuring the belief remains current as new observations are made.

\subsection{State Estimation Network}
Classical Bayesian state estimation is impractical in the learned state space due to unknown motion and observation models, as well as the potential for non-Markovian effects. To address these challenges, \xmobility{} introduces a state estimation network, where a history state $h_t$ is used to capture non-Markovian effects, together with a recurrent unit to propagate the history state over time. With the observation embeddings $o_t$ produced by observation encoders and the applied action $a_{t}$, the belief state is modeled as:
\begin{equation}
s_t\sim \mathcal{N}(\mu_\theta^{e}(h_{t-1},a_{t-1}, o_{t}), \sigma_\theta^{e}(h_{t-1},a_{t-1}, o_{t})\mathit{I}),
\label{eqn:state_estimation}
\end{equation}
where the recurrent history transition follows:
\begin{equation}
h_{t}=f_{\theta}(h_{t_{t-1}}, s_{t}).
\end{equation}
The history $h_{t-1}$ and state $s_t$ are then concatenated to form a 1-D latent state $z_t = [h_{t-1}, s_t]$ used for multi-task decoding.

\subsubsection{Model Implementation}
To model these transitions, the $f_{\theta}$ is implemented as a gated recurrent unit (GRU), while $(\mu_\theta^{e}, \sigma_\theta^{e})$ are modeled as multi-layer perceptrons (MLPs). More specifically, within the state estimator as shown in Fig. \ref{fig:model_arch}(b), the input action command is first processed through an MLP to produce a higher-dimensional feature state. This feature state is then concatenated with the history and observation embedding to estimate $(\mu_\theta, \sigma_\theta)$ via a normal distribution network. Finally, a sampler draws a state from the learned distribution.

\subsubsection{Observation Encoding}
To update the belief state, \xmobility{} takes front-view image and robot states, including linear forward velocity, as inputs. These inputs are embedded into a compact, low-dimensional representation to effectively learn the dynamics described in Eqn. (\ref{eqn:state_estimation}). To capture visual features, we employ a pretrained DINOv2 \cite{oquab2023dinov2} model as the image encoder. The image is embedded into a 1-D token $u_t \in \mathbb{R}^{768}$ by concatenating the class token with the average-pooled patch tokens. DINOv2's strong feature representation and robust performance across various image conditions can enhance \xmobility{}'s ability to generalize well across different environments in both simulation and real world. The linear forward speed is encoded into $m_t \in \mathbb{R}^{32}$ using fully connected layers. This speed embedding is then concatenated with the image embedding to produce the observation embedding $o_t = [u_t, m_t]$ for state estimation.

\subsubsection{Multi-task Decoding}
Making sure the latent state $z_t$ captures necessary information is essential for the downstream navigation skill training, which requires well-designed decoders. In \xmobility{}, we adopt two decoding heads: RGB reconstruction and semantic segmentation as in Fig. \ref{fig:model_arch}(e). For RGB reconstruction, we fine-tuned a latent diffusion model (LDM) \cite{rombach2021highresolution}, where the latent state $z_t$ conditions the UNet during the denoising process, following a similar approach as in \cite{wang2023drivedreamer}. The difference between the imposed and predicted noises is penalized via a MSE loss. Semantic segmentation is conducted in a perspective view to more accurately capture stacked obstacles that are not well represented in a bird’s-eye view, a method widely used in the autonomous driving domain \cite{Yang2022BEVFormerVA}. To achieve this, we utilize StyleGAN \cite{karras2019style} to decode the semantic segmentation from the latent state, starting with a learned constant tensor that is progressively upsampled to the final resolution. At each upsample resolution, we compute the Cross-Entropy loss for the predicted semantics .

By training the encoders, decoders, and state estimation network together end-to-end, \xmobility{} can produce a rich latent state that effectively supports action policy learning.

\subsection{State Prediction Network}
To support applications such as long-horizon planning, which require predicting sequences of planned actions, \xmobility{} also introduces a state prediction network that forecasts the consequences of future actions. Without using observation embeddings, the belief state in the state predictor is modeled as:
\begin{equation}
s_t\sim \mathcal{N}(\mu_\theta^{p}(h_{t-1},a_{t-1}), \sigma_\theta^{p}(h_{t-1},a_{t-1})\mathit{I}).
\label{eqn:state_prediction}
\end{equation}
The state prediction network is implemented similarly to the state estimation network, with a GRU for recurrent history updates and a Normal distribution model for the state, but without observation embeddings as input.

For state predictor training, a Kullback-Leibler divergence loss is applied between the state estimator and state predictor's belief distributions, encouraging the state predictor to match the state estimator. This allows the model to predict future states that match observed data, improving navigation performance in dynamic environments and enabling broader applications of the world model in various contexts \cite{wang2023drivewm}.

\subsection{Action Policy Network}
With the world modeling in place, the latent state can provide a rich representation of both the environment and the robot's state. The action policy network, depicted in Fig. \ref{fig:model_arch}(e), leverages this latent state $z_t$, along with the route encoding $g_t$, to predict the action commands $a_t \sim \pi(a_t|z_t, g_t)$.

\subsubsection{Route Encoding}
The route provides essential high-level guidance for navigation, particularly in large-scale environments. While robot localization is required for \xmobility{}, the route can be as simple as a straight line from the robot's position to its destination, or it can consist of a sequence of waypoints from the environment's topological graph, if available. This design enables \xmobility{} to function effectively in large environments while avoiding the costly route search used in \cite{macenski2020marathon2}.

To digest route information, we first transform the global route into the robot's frame, allowing the model to operate without requiring a map. We then truncate the regional route segment near the robot, obtaining a series of route poses with x and y positions. This segment is encoded using VectorNet \cite{gao2020vectornet}, which produces a 1-D feature embedding $g \in \mathbb{R}^{64}$. Compared to rasterizing the route as an image \cite{hu2022model}, VectorNet can more effectively capture route details and offer the flexibility to encode additional attributes, such as the final destination flag, which are critical for robot navigation.

\subsubsection{Action Decoding}
To decode action policies, the latent state and route embedding are fused using a self-attention mechanism, allowing the model to balance route adherence with environmental interaction. The fused policy state $p_t$ is then decoded by an MLP to generate action commands $a \in \mathbb{R}^{6}$, representing the desired linear and angular velocities in the x, y, and z directions, along with an optional navigation path $p \in \mathbb{R}^{5\times2}$, which includes five path poses in the robot frame. The action policy head is trained using an $L1$ loss to imitate the teacher policy. For state estimation or prediction, only the action commands are used, while the imitated path can better shape the latent states and support post-process tasks like safety checking.

\subsection{Multi-stage Training}
As illustrated in Fig. \ref{fig:model_arch}, the action $a_{t}$ can originate either from the learned policy or from the dataset. This flexibility allows for a multi-stage training pipeline. In the first stage, we deactivate the action policy network and use logged actions from the off-policy dataset to train the world model. Once the world model is sufficiently trained, we activate the action policy network and train it alongside the world model using the on-policy dataset in the second stage. This approach efficiently utilizes large-scale datasets that lack high-quality teacher policies and allows the model to fully explore world dynamics. This is one of the key factors that enables \xmobility{} to generalize effectively across out-of-distribution environments not present in the training dataset.

\section{Experiment Setting}\label{section:exp_setting}

\subsection{Dataset}
The training data was collected using the Nova Carter robot in Isaac Sim. This photorealistic synthetic dataset includes two types of action policy inputs: random actions and a teacher policy based on Nav2 \cite{macenski2020marathon2}. For the random action dataset, a customized Synthetic Data Generation (SDG) pipeline is built in Isaac Sim with modularity design in mind. During the data collection, we first randomly sample robot initial positions, and then apply random actions for a number of steps before we switch to a new set of random actions. A carefully designed randomization logic is implemented to ensure comprehensive state-action coverage, therefore allowing enough exploration of the environment. For the teacher policy dataset, we integrated Nav2 into this pipeline for a closed-loop simulation, with navigation tasks generated by randomly sampling start and goal pairs.

\begin{figure}[t]
\begin{center}
\centering\includegraphics[width=3.4in]{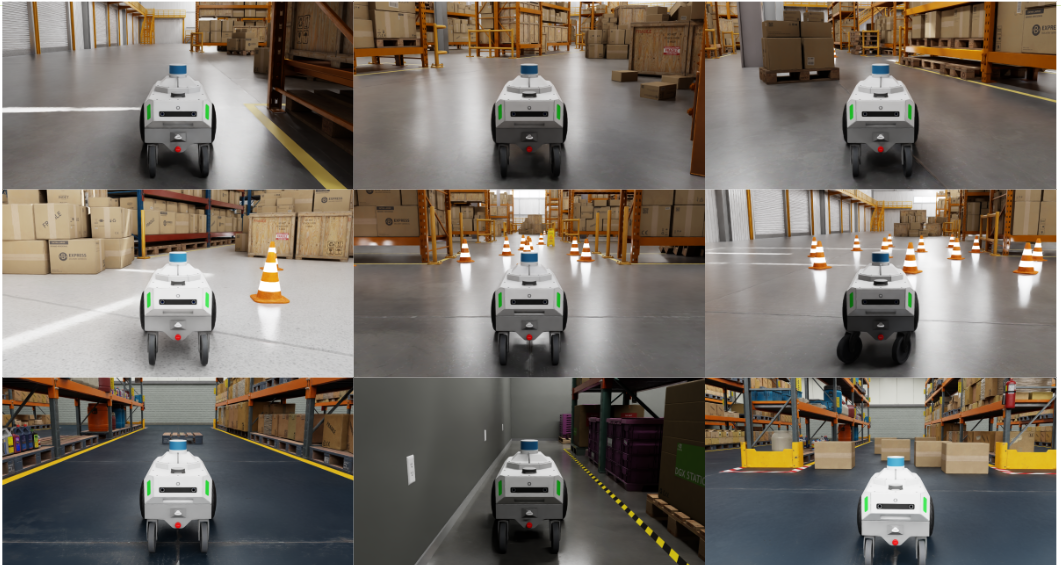}
\caption{Benchmark warehouse scenarios with varying levels of difficulty.}
\label{fig:benchmark_scenarios}
\end{center}
\end{figure}

Data was recorded at 5Hz across four distinct warehouse environments, each featuring randomly generated textures and layouts. The lighting diversity is also kept as realistic as possible, which can be seen in Fig. \ref{fig:benchmark_scenarios}. Warehouses were chosen because they provide large-scale, semi-structured, and cluttered settings, offering a diverse range of scenarios for the dataset. In total, we gathered 100K frames from the Nav2 policy and 160K frames from the random action policy, which are evenly distributed across the four scenarios. Each frame includes key fields as outlined in Table \ref{tab:dataset_item}. The \textit{image} field contains the RGB input from the front-view camera, while the \textit{semantic label} field identifies each pixel according to the predefined semantic classes: [\textit{Navigable, Forklift, Cone, Sign, Pallet, Fence, Background}], which is to cover critical object types for safe navigation in warehouse environments. The \textit{route} field contains the regional route segment transformed into the robot’s frame, and the \textit{speed} field captures the robot's linear and angular velocities. The learning targets,  \textit{path} and \textit{action command}, include a sequence of path poses and the desired linear and angular velocities in the x, y, and z dimensions.

\begin{table}[t]
\centering
\begin{threeparttable}
\caption{Dataset Elements}
\begin{tabular}{c l c c}
\toprule
Field & Shape & Teacher & Random Action \\
\midrule
\textbf{image} & $\mathbb{R}^{ 3 \times 320 \times 512}$ & $\checkmark$ & $\checkmark$ \\
\textbf{speed} & $\mathbb{R}^{1}$ & $\checkmark$ & $\checkmark$ \\
\textbf{semantic label} & $\mathbb{R}^{ 7 \times 320 \times 512}$ & $\checkmark$ & $\checkmark$ \\
\textbf{route} & $\mathbb{R}^{20\times2}$ & $\checkmark$ & $\times$  \\
\textbf{path} & $\mathbb{R}^{5\times2}$ & $\checkmark$ & $\times$\\
\textbf{action command} & $\mathbb{R}^{6}$ & $\checkmark$ & $\checkmark$ \\
\bottomrule
\end{tabular}
\label{tab:dataset_item}
\end{threeparttable}
\end{table}

\subsection{Training}
The world model was first trained for 100 epochs using the random action dataset, with a batch size of 32, distributed across 8 H100 GPUs. Each training sample contained 5 frames of data to facilitate world dynamics learning in an auto-regressive manner. Following this, the action policy was trained alongside the world model for an additional 100 epochs using the teacher policy dataset. The RGB diffuser was disabled in the action policy stage to improve training speed. The AdamW optimizer was employed with a learning rate of $10^{-5}$. 

\subsection{Evaluation}

The model was evaluated in both open-loop and closed-loop settings. For the open-loop evaluation, we measured the Mean Absolute Error (MAE) for linear speed (L-MAE), angular speed (A-MAE), and path prediction (P-MAE) using the test split of the training dataset. In the closed-loop evaluation, we built a navigation benchmark suite with 10 warehouse scenarios of varying difficulty, including narrow corridors, cluttered spaces, and low-lying obstacles (see Fig. \ref{fig:benchmark_scenarios}). To assess the model’s performance in out-of-distribution cases, 5 of these warehouse scenarios were set in environments not included in the training data. Additionally, 50 randomly generated cluttered scenes in narrow corridors were created to systematically evaluate generalization capability. Each method was tested over 100 runs by conducting 5 trials per designed warehouse scenario. We tracked three key metrics: 1) mission success rate (SR), 2) weighted trip time (WTT), which measures navigation efficiency by dividing trip time by success rate, and 3) average absolute angular acceleration (AA) as an indicator of motion smoothness.

\section{Results} \label{section:results}

\subsection{Navigation Performance} 
We compared \xmobility{}'s navigation performance against several baselines: its teacher policy Nav2, a model-free behavior cloning (BC) method, and a prior state-of-the-art model-based imitation learning method, MILE \cite{hu2022model}. The BC method used the same observation encoder as \xmobility{} but directly mapped observations to policies. For MILE, we adapted the model to predict semantic information in a perspective view rather than a bird's-eye view (BEV), as originally proposed, to better align with robotic applications. All methods were retrained on the same dataset to ensure a fair comparison.

\begin{table}[t]
\centering
\caption{Open Loop Metrics}
\begin{threeparttable}
\begin{tabular}{l | c c c}
\toprule
 Method & \textbf{A-MAE} $\downarrow$ & \textbf{L-MAE} $\downarrow$ & \textbf{P-MAE} $\downarrow$    \\ 
\midrule
MILE   & 0.0751 & 0.0106 & 0.1192        \\ 
BC  & 0.0870 & 0.0165 & 0.1176 \\
\xmobility{} &  0.0747 & 0.0158 & 0.1156\\
\xmobility{} w/ DP\tnote{*} &  0.0306 & 0.1115 & 0.0389\\
\bottomrule
\end{tabular}
\begin{tablenotes}
\footnotesize
\item[*] DP: Diffusion Policy
\end{tablenotes}
\label{tab:open_loop_metrics}
\end{threeparttable}
\end{table}

\begin{table*}[ht]
\centering
\caption{Close Loop Navigation Benchmark}
\begin{tabular}{l | c c c  | c c c }
\toprule
  & \multicolumn{3}{c|}{\textbf{Warehouse Scenarios}} & \multicolumn{3}{c}{\textbf{Random Obstacle Scenarios}}    \\ 
\textbf{Method} &  \textbf{SR(\%)} $\uparrow$ & \textbf{WTT(s)} $\downarrow$ & \textbf{AA($rad/s^{2}$)} $\downarrow$ & \textbf{SR(\%)} $\uparrow$& \textbf{WTT(s)} $\downarrow$ & \textbf{AA($rad/s^{2}$)} $\downarrow$     \\ 
\midrule
ROS Nav2                  & 58 & 68.41  & 0.606 & 34  & 74.94 & 0.391     \\ 
MILE                      & 34 & 128.32 & 0.274 & 26 & 113.49 &  0.212       \\ 
BC                        & 92 & 40.4   & 0.219 & 56 & 42.72 & 0.219  \\
\xmobility{}  & \textbf{96} & \textbf{37.7}  & \textbf{0.206} & \textbf{68} & \textbf{40.21} & \textbf{0.186} \\
\midrule
\multicolumn{7}{l}{\textbf{Ablation Studies}} \\
\midrule
\xmobility{} w/o Pretrain  & 88 &  43.97   & 0.215 & 44 & 65.45 & 0.213 \\
\xmobility{} w/o Semantic & \textbf{96} & \textbf{33.7}  & \textbf{0.203} & 36 & 77.77 & 0.273 \\
\xmobility{} w/o History Tracking & 72 & 37.5 & 0.264 & 16 & 175.10 & 0.229  \\  
\xmobility{}  & \textbf{96} & 37.7  & 0.206 & \textbf{68} & \textbf{40.21} & \textbf{0.186} \\
\bottomrule
\end{tabular}
\label{tab:nav_stack_benchmark}
\end{table*}

As shown in Table \ref{tab:open_loop_metrics}, all learning-based methods were properly trained and achieved similar open-loop metrics on the test dataset. In the close loop evaluation, \xmobility{} consistently outperforms the other methods across all navigation metrics, as detailed in Table \ref{tab:nav_stack_benchmark}. Compared to it's teacher Nav2, \xmobility{} achieves higher mission success rates and smoother motion in every scenario. In tests involving narrow corridors and multiple homotopy classes, Nav2 often fails or exhibits indecision, exposing common limitations of classical approaches. In contrast, \xmobility{} successfully navigates these challenges by reasoning through key navigation skills learned from demonstrations and producing smoother motion via leveraging its ability of retaining a history of actions and observations. Behavior Cloning (BC), which lacks the world dynamics understanding and history tracking, results in less smooth trajectories and lower success rates, particularly in out-of-distribution random obstacle scenarios. MILE performs the worst, with the robot struggling to follow the designated route, likely due to route information loss during encoding and feature compression. Additionally, MILE's reliance on ResNet-18 for image encoding, instead of DINOv2, may further contribute to its underperformance.

\subsection{Ablation Studies}

\subsubsection{World Model Pretrain}
Pretraining the world model on the 160K frames of random action data leads to improvements in navigation performance, particularly in scenarios involving random obstacles. This demonstrates the model's enhanced ability to manage out-of-distribution cases by leveraging off-policy data. Additionally, semantic segmentation performance improves as well, benefiting from the expanded dataset used for training the semantic decoder, as shown in Table \ref{tab:semantic_seg}.

\begingroup
\setlength{\tabcolsep}{4pt} 
\begin{table}[t]
\centering
\caption{Semantic Segmentation IOU}
\begin{tabular}{l c c c c c c}
\toprule
Method & Navigable & Forklift & Cone & Pallet & Fence & Sign \\
\midrule
MILE & 0.96& 0.08 &0.09 & 0.26  & 0.19& 0.07 \\
 w/o Pretrain & 0.97 & 0.13 & 0.14 & 0.38 & 0.27 & 0.12\\
\textbf{\xmobility{}} &\textbf{ 0.97} & \textbf{0.16 }& \textbf{0.18 }& \textbf{0.42 }& \textbf{0.32} & \textbf{0.15}\\
\bottomrule
\end{tabular}
\label{tab:semantic_seg}
\end{table}
\endgroup

\subsubsection{Semantic Decoding}
\begin{figure}[t]
\begin{center}
\centering\includegraphics[width=3.4in]{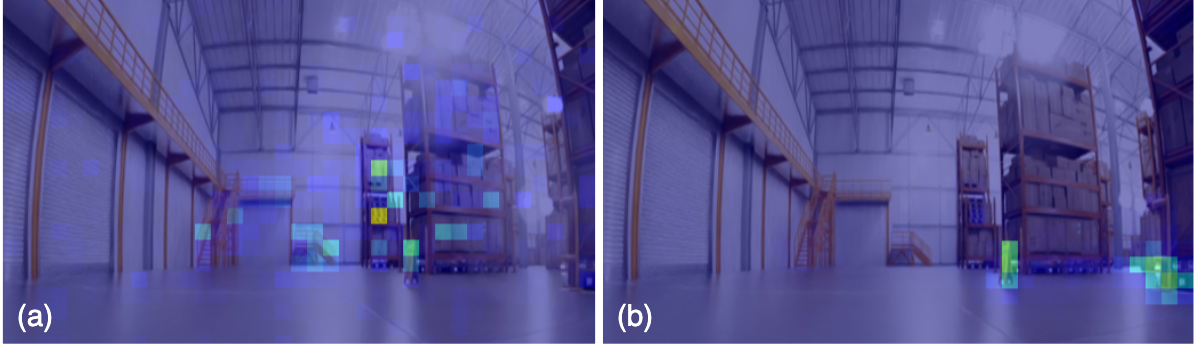}
\caption{Attention analysis: (a) pretrained DINOv2, (b) fine tuned with semantic decoding. Attention is directed toward key semantic objects (e.g., sign, pallet, fence) when semantic decoding is enabled, as opposed to being scattered across the entire image without semantic decoding. }
\label{fig:attention_analysis}
\end{center}
\end{figure}

To assess the impact of semantic decoding on action policy, we compared methods with and without semantic segmentation, as shown in Table \ref{tab:nav_stack_benchmark}. Without semantic decoding, the model performs better in warehouse scenarios but shows a decline in performance in random obstacle environments, indicating reduced generalizability. This supports our hypothesis that semantic decoding plays a important role in learning world dynamics, ensuring that the latent state contains meaningful environment information for navigation, which is useful for effective policy learning. This conclusion is further reinforced by the attention plots in Fig. \ref{fig:attention_analysis}, where the attention is correctly directed toward key semantic objects when semantic decoder is enabled. The improved performance in warehouse scenarios without semantic decoding suggests possible overfitting to the specific environments in the dataset.

\subsubsection{History Tracking}
To assess the impact of history tracking strategies, we compared the closed-loop performance of our model using two approaches: resetting and fully recurrent. In the resetting approach, the history and latent state are reinitialized at the start of each cycle, while in the recurrent mode, the history and latent state are initialized only once at the beginning and continuously updated with new observations. The recurrent mode showed superior performance, resulting in much smoother motion. In contrast, the resetting mode led to issues such as the robot frequently overshooting during turns. These results indicate that the model effectively learns a representation capable of integrating history over longer horizons, beyond those encountered during training.

\subsubsection{Diffusion Policy}
To enhance the action policy, we also experimented with the diffusion policy \cite{chi2024diffusionpolicy}, conditioning on policy state $p_t$ to jointly decode action commands and paths. The diffusion policy produced mixed results overall. While path prediction is improved, the linear action commands degraded notably, as reflected in the open loop metrics in Table \ref{tab:open_loop_metrics}. In closed-loop evaluation, the diffusion policy struggled to provide consistent action commands and had difficulty in navigating to the destination without collisions. A possible explanation is that the model generates single-step commands instead of predicting a sequence, leading to a low signal-to-noise ratio during the denoising process, then causes instability. Further investigation into this issue is planned as future work.

\subsection{World State Prediction}

\begin{figure}[t]
\begin{center}
\centering\includegraphics[width=3.4in]{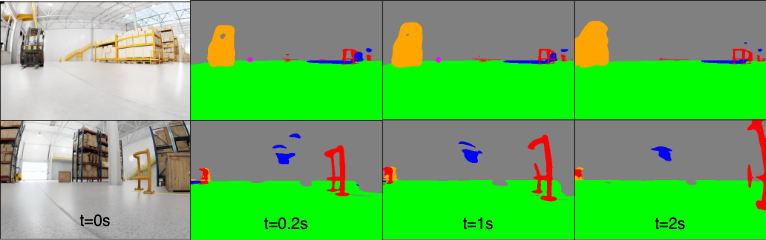}
\caption{Qualitative examples of prediction by decoding semantic segmentation from latent state. Green: Navigable, Red: Fence, Blue: Pallet, Orange: Forklift, Purple: Sign}
\label{fig:semantic_prediction}
\end{center}
\end{figure}

\begin{figure}[t]
\begin{center}
\centering\includegraphics[width=3.4in]{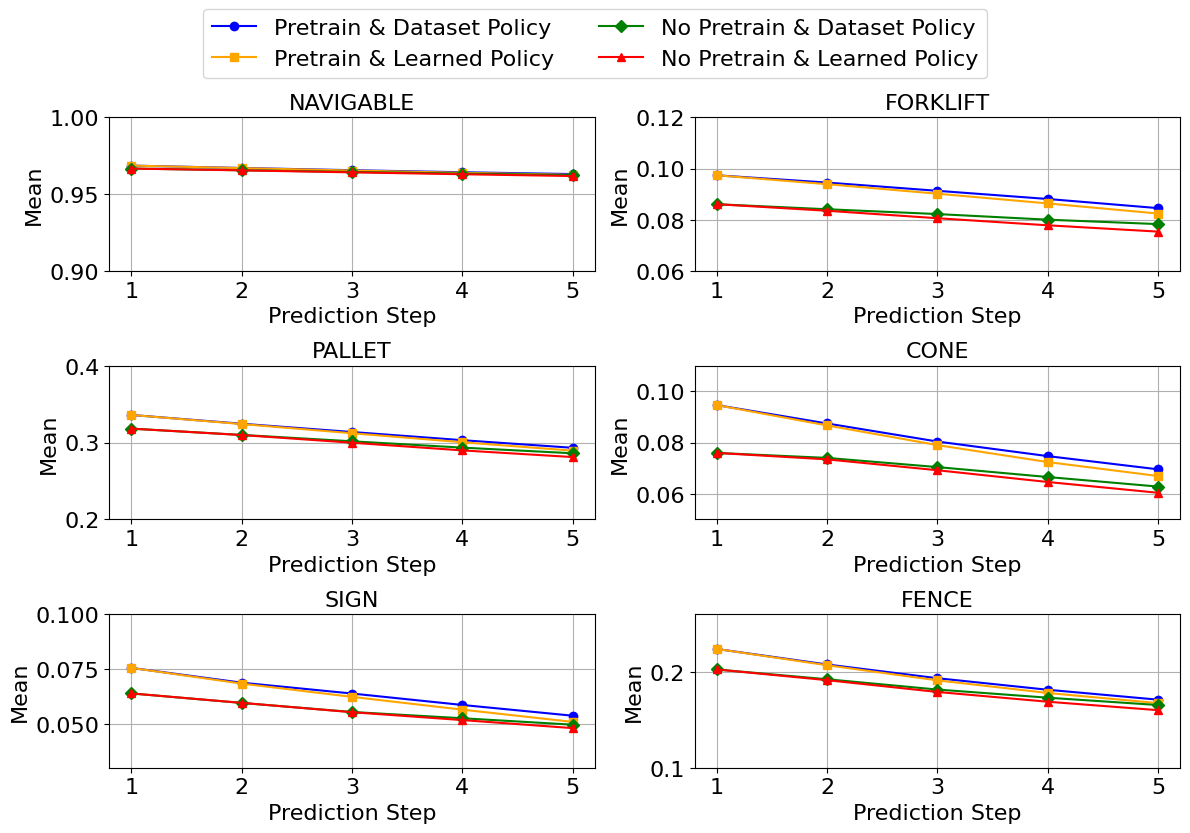}
\caption{IOU for semantic decoding at predicted steps.}
\label{fig:semantic_iou}
\end{center}
\end{figure}

With state predictor, \xmobility{} can predict future states in the latent space, which are then decoded into semantic segmentation for better interpretability, as shown in Fig. \ref{fig:semantic_prediction}. To evaluate the quality of these predictions, we calculated the Intersection over Union (IOU) for the predicted semantic segmentations over 5 predicted steps. The results, shown in Fig. \ref{fig:semantic_iou}, compare \xmobility{} with and without pretraining, and with or without the learned policy enabled for prediction. Across all setups, the model can consistently detect and predict the navigable surface, a key element for successful navigation. \xmobility{} without pretraining shows lower IOU across all semantic classes compared to its pretrained counterpart. As expected, enabling the learned action policy introduces more prediction error as the prediction horizon extends.

\section{Discussions} \label{section:discussion}

\subsection{Sim2Real Transfer}
We successfully deployed \xmobility{} on a Nova Carter robot without any fine-tuning, demonstrating its navigation capabilities through zero-shot Sim2Real transfer. The robot was able to safely navigate through cluttered obstacles in a lab environment (see Fig. \ref{fig:x_mobility}), despite the environment being absent from the training dataset, further showcasing its ability to generalize to out-of-distribution scenarios. 

\begin{figure}
\begin{center}
\centering\includegraphics[width=3.4in]{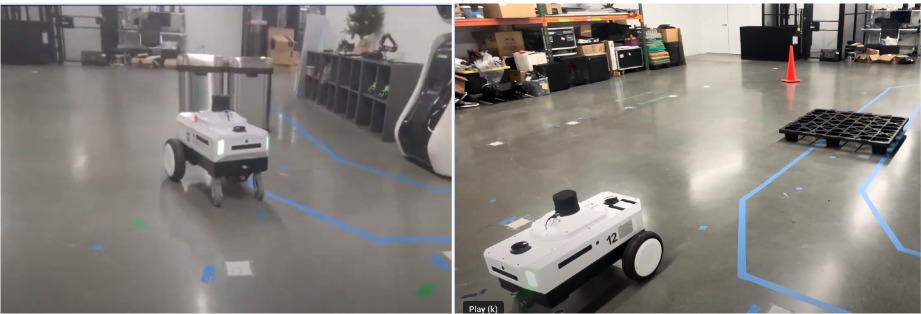}
\caption{Environments setup for benchmark on real robot.}
\label{fig:real_env}
\end{center}
\end{figure}

To more rigorously evaluate real-world performance, we set up environments with varying obstacle configurations and lighting conditions (see Fig. \ref{fig:real_env}). The final results, summarized in Table \ref{tab:real_env}, demonstrate performance comparable to the benchmarks observed in simulation. Additionally, we also benchmarked the model's run-time performance on the Jetson AGX Orin,  as showing in Table \ref{tab:trt_run_time}, underscoring \xmobility{}'s high computational efficiency for edge-device navigation.

\begin{table}
\centering
\caption{Benchmark on real robot}
\begin{tabular}{c c c c c}
\toprule
\textbf{Scene} & \textbf{Lightning} & \textbf{SR} & \textbf{AA($rad/s^2$)} \\ 
\midrule
Single Obstacle & Normal & 10/10 & 0.267 \\ 
Single Obstacle & Dark & 10/10 & 0.267 \\ 
Multi Obstacles & Normal & 8/9 & 0.205 \\ 
Multi Obstacles & Dark & 8/9 & 0.273 \\ 
\bottomrule
\end{tabular}
\label{tab:real_env}
\end{table}

\begin{table}[t]
\centering
\caption{Inference Latency and GPU Memory Usage}
\begin{tabular}{c c c c c}
\toprule
\textbf{Inference Mode} & \textbf{P50(ms)} & \textbf{P95(ms)} & \textbf{GPU Memory(MB)} \\ 
\midrule
\textbf{Policy Only} & 38.6 & 41.96 & 594 \\ 
\textbf{Policy + Semantic} & 55.55 & 55.72 & 804 \\ 
\bottomrule
\end{tabular}
\label{tab:trt_run_time}
\end{table}

We believe that several factors contribute to the zero-shot Sim2Real transferability. First, the DINOv2 encoder offers strong feature extraction and representation for input images under various conditions. Second, the probabilistic world model design can enhance the system robustness, particularly against dynamic noises from the real robot, which cannot be fully captured during simulation. Lastly, the synthetic dataset collected using Isaac Sim's Digital Twin is highly photorealistic, further enabling smooth Sim2Real transfer.

\subsection{Cross Embodiment}

\begin{figure}[t]
\begin{center}
\centering\includegraphics[width=3.4in]{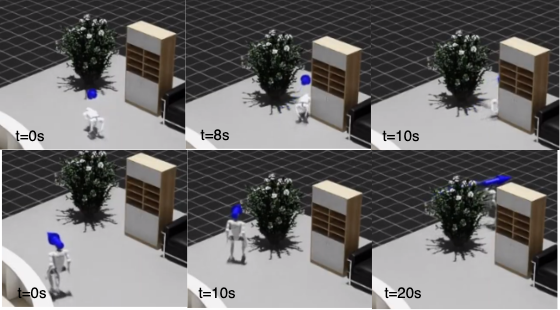}
\caption{\xmobility{} on humanoid and quadruped robots with RL-based locomotion policy pre-trained in Isaac Lab.}
\label{fig:humanoid}
\end{center}
\end{figure}

To further evaluate \xmobility{}'s generalization capability, we also tested its performance across different embodiments. Thanks to its end-to-end design and standardized input and output formats, we successfully deployed \xmobility{} on four robot platforms in Isaac Sim: Nova Carter (differential drive), Forklift (Ackermann drive), Unitree Go2 (quadruped), and Unitree G1 (humanoid). By integrating \xmobility{} with the appropriate controllers or locomotion policies pre-trained in Isaac Lab \cite{mittal2023orbit}, each robot was able to navigate successfully (see Fig. \ref{fig:humanoid}). While navigation performance varied depending on camera placement and the robots' dynamic constraints, \xmobility{} consistently demonstrated good potential for generalization across different embodiments.

\section{Future Work}\label{section:future_work}
In this work, we proposed \xmobility{}, a generalizable navigation model that integrates world modeling with imitation learning to enhance navigation performance and generalization across out-of-distribution scenarios.

Looking ahead, future work will focus on further enhancing \xmobility{}'s cross-embodiment capability. This will involve incorporating more detailed robot specification encoding to better adapt the model to different platforms. Additionally, we plan to leverage RL fine-tuning to refine the model's performance across various embodiments. Another key direction will be to enrich the dataset with more diverse scenes with dynamic obstacles, enabling a deeper exploration of the world model's role in supporting action policy learning. These advancements will help in solidifying \xmobility{} as a robust and adaptable solution for navigation across a wide range of environments and robot types.

\bibliographystyle{IEEEtran}
\bibliography{reference}


\appendices
\section{Model Description}
The model parameters of  \xmobility{} are listed in Table \ref{tab:model_parameters}, while Table \ref{tab:hyperparameters} outlines the hyperparameters used for training.

\begin{table}[ht]
\centering
\caption{Model parameters}
\begin{tabular}{c l r }
\toprule
 & \textbf{Name} & \textbf{Parameters} \\ 
\midrule
\multirow{2}{*}{Encoders} & Image encoder & 22.1M \\ 
 & Robot state encoder & 1.1K \\
 \midrule
\multirow{2}{*}{State Estimators} & State estimator & 5.5M \\
 & State predictor & 2.3M \\
 \midrule
\multirow{4}{*}{Action Policy} & Route encoder & 8.5K \\
 & Self-attention fuser & 28.5M \\
 & Action command decoder & 10.5M \\
 & Path decoder & 10.5M \\
 \midrule
\multirow{2}{*}{Decoders}  & Semantic decoder & 34.2M \\
 & RGB diffuser & 962M \\
\bottomrule
\end{tabular}
\label{tab:model_parameters}
\end{table}

\begin{table}[h]
\centering
\caption{Hyperparameters}
\begin{tabular}{c l r }
\toprule
\textbf{Category} & \textbf{Name} & \textbf{Value} \\ 
\midrule
\multirow{5}{*}{Training} & GPUs & 8 H100 \\
 & Batch size & 32 \\ 
 & Precision & 16-mixed \\
 & World model epoches & 100 \\
 & Action policy epoches & 100 \\
 \midrule
\multirow{5}{*}{Optimizer}  & Type & AdamW \\
 & Learning rate & 1e-5 \\
 & Weight decay & 0.01 \\
 & Scheduler & OneCycleLR \\
 & Scheduler pct start & 0.2 \\
 \midrule
\multirow{3}{*}{Observation Encoders} & Image size & $320\times512$ \\
& Image embedding dim & 768 \\
 & Robot state embedding dim & 32 \\
 \midrule
\multirow{4}{*}{State Estimation} & History dim & 1024 \\
 & State dim & 512 \\
 & Action encoding dim & 64 \\
 \midrule
\multirow{2}{*}{Action Policy}& VectorNet layers & 4 \\
 & Policy state dim & 2048 \\ 
 \midrule
\multirow{1}{*}{Semantic Decoder} & StyleGan constant size & (5, 8) \\
\midrule
\multirow{6}{*}{RGB Diffuser} & Noise scheduler & LMS discrete \\
& Beta schedule & scaled linear \\
& Beta start & 0.00085 \\
& Beta end & 0.012 \\
& Training timesteps & 1000 \\
& Inference timesteps & 50 \\
\midrule
\multirow{6}{*}{Losses} & Action command weight & 10.0 \\
& Path weight & 5.0 \\
& Semantic weight & 1.0 \\
& RGB weight & 10.0 \\
& KL weight & 0.001 \\
& KL balancing alpha & 0.75 \\
\bottomrule
\end{tabular}
\label{tab:hyperparameters}
\end{table}

\end{document}